\documentclass[12pt]{iopart}
\usepackage{graphicx} 
\usepackage{soul}
\PassOptionsToPackage{hyphens}{url}\usepackage{hyperref}

\usepackage{xcolor}
\usepackage{csquotes}

\begin{document}

\title[]{Outlook Towards Deployable Continual Learning for Particle Accelerators}

\author{Kishansingh Rajput$^{*, 1, 2}$, Sen Lin$^{2}$, Auralee Edelen$^{3}$, Willem Blokland$^{4}$, Malachi Schram$^{1}$}
\address{$^1$ Thomas Jefferson National Accelerator Facility, Newport News, VA 23606, USA}
\address{$^2$ Department of Computer Science, University of Houston, Houston, TX 77204, USA}
\address{$^3$ SLAC National  Laboratory, Menlo Park, California 94025, USA}
\address{$^4$ Spallation Neutron Source, Oak Ridge National Laboratory, Oak Ridge, TN, USA}
\ead{$^*$ kishan@jlab.org}

\vspace{10pt}
\begin{indented}
\item[]February 2025
\end{indented}

\begin{abstract}
Particle Accelerators are high power complex machines.
To ensure uninterrupted operation of these machines,
thousands of pieces of equipment need to be synchronized, which requires addressing many challenges including design, optimization and control, anomaly detection and machine protection. 
With recent advancements, Machine Learning (ML) holds promise to assist in more advance prognostics, optimization, and control. While ML based solutions have been developed for several applications in particle accelerators, only few have reached deployment and even fewer to long term usage, due to particle accelerator data distribution drifts caused by changes in both  measurable and non-measurable parameters. 
In this paper, we identify some of the key areas within particle accelerators where continual learning can allow maintenance of ML model performance with distribution drifts. Particularly,
we first discuss existing applications of ML in particle accelerators, and their limitations due to distribution drift.
Next, we review existing continual learning techniques and investigate their potential applications to address data distribution drifts in accelerators. By identifying the opportunities and challenges in applying continual learning, this paper seeks to open up the new field and inspire more research efforts towards deployable continual learning for particle accelerators.

\end{abstract}

\newpage
\section{Introduction}

Particle accelerators~\cite{particle_accelerator} are of great importance in the investigation of the fundamental building blocks of matter and advancing our understanding of the universe with many facilities globally operating them to conduct science experiments. 
Smaller scale particle accelerators are also leveraged in many industrial applications, including sanitation, medical imaging and cancer therapy~\cite{Peach2011}.
To maintain optimal performance, these accelerator systems 
require meticulous tuning and adjustment. Typically, operators utilize 
either manual fine-tuning by adjusting many precision knobs or 
leverage sophisticated automated optimization algorithms to achieve 
precise control.

This is very challenging due to the large number of adjustable parameters and the sensitivity of the beam responses. 
As such, operators spend significant effort and time tuning the accelerator beam at the beginning of a run, especially if the accelerator was down for a long time or any upgrades to the equipment had been performed. 
In some cases there is a lack of appropriate physics models to capture the behavior of the accelerated beam, such as beam halo losses~\cite{751035, Nghiem2012BEAMHD}. 
A "beam halo" refers to the region around the 
desired particle beam where particles are being deflected or scattered due 
to various sources of radiation and interactions with surrounding 
materials.
Heuristic approaches, e.g. operators manually adjusting the accelerator settings to minimize losses, are used which often require significant time to complete the optimization. 
These challenges can be exacerbated if an accelerator needs to frequently switch among
different types of beams or beam power requiring different settings for the equipment. 
For example, the Facility for Rare Isotope Beams (FRIB) uses different beam species for different experiments. 
Adjustments to control algorithms must be made and/or the heuristic knowledge must be learned for the new setup. 

During operation, the accelerator beam can easily stray from the intended path when any equipment deviates from its expected behavior, resulting in radio-activation or worse, damage, unless the beam is quickly aborted.  
This is a problem especially for high-power beams, where even stray halo can cause damage. 
While preventative maintenance is scheduled and many diagnostics sensors are placed along the beam line to detect these errant beam conditions and protect the equipment from damage by aborting the beam, unexpected downtime due to faults still occur.

Additional complexity arises from the tendency of accelerators to experience both gradual and systematic change in behavior or 
characteristics over time during operation due to various reasons including machine aging, and environmental factors.
This phenomenon, referred as \textquote{drift}, can significantly impact performance and 
requires adjustments in accelerator settings.
As such, there are challenges with respect to:
1) minimizing time spent on tuning for different setups to maximize time for experiments,
2) optimizing performance in the absence of physics models, and
3) detecting and minimizing the impact of abnormal or drifting conditions. 

With the rapid increase in the storage and computational power, Machine Learning (ML)  has achieved astonishing successes in  addressing complex problems in science and engineering, including material science~\cite{ML_MaterialScience}, bio-informatics~\cite{ML_Bioinformatics}, radiology~\cite{ML_radiology}, renewable energy~\cite{ML_RenewableEnergy}, physics~\cite{ML_Physics, Boehnlein_2022} and many other fields. 
In particular, 
ML-based methods can outperform many traditional methods by leveraging 
Artificial Neural Networks (ANNs)~\cite{ANN} for feature learning. Thus motivated, there have emerged a lot of studies to develop
ML-based solutions for various applications in particle accelerators, including anomaly detection and prediction \cite{Rajput_2024, Blokland:2021onk, edelen_anomaly_2021, ALANAZI2023100484, 9658806}, optimization \cite{osti_761286, rajput2024DDRL}, dynamic control \cite{EdelenTNS2016, MeierNIM2009, EdelenNeurIPS2017}, and surrogate models \cite{Kafkes:2021jse, MCSPADDEN2024100518, schram2023uncertainty, LOI_ML, LOI_ML2, Gupta_2021, ogren2021surrogate}. ML techniques have also been adopted in 
real-time decision support and control applications for more accurate and faster turn-around~\cite{real-time_ML, real-time_ML2, PhysRevAccelBeams.24.104601}. 

Nevertheless, the evolving nature of real-world systems poses a significant challenge for current ML approaches being directly used in practice: standard ML assumes a stationary distribution for the data, whereas the data distribution and the underlying concept can dynamically change with time in real systems.
In dynamic machines such as particle accelerators, both data and the input to target relationship drift over time due to various reasons~\cite{Alex-data_drift_compact, Auralee-data-drift-NN}, which may lead to a substantial performance drop of a deployed ML model~\cite{Concept_drift}. 
Therefore, a critical requirement towards deployable ML is the model robustness in predictions, especially against inherent noise in accelerator measurements, occasional failures of signals (e.g., a read-back failure), and both deliberate and unintended time-varying changes in conditions.
These can result in new data patterns outside of the statistical distribution of the training data \cite{Concept_drift}, commonly known as concept drift~\cite{Lu_2018},
which is difficult for ML models to handle due to their limited capacity to extrapolate beyond the scope of the training distribution \cite{AI-aging}. In particular, 
concept shift is more evident in particle accelerators due to continuous stream of drifting data, and developing robust learning systems is essential to carry out the practical benefits of ML in this domain.

Continual learning \cite{PARISI201954, lopezpaz2022gradient, wang2024comprehensive, schwarz2018progress}, also known as lifelong learning, addresses performance degradation with continual model updates  under data distribution shifts.
It provides a promising solution for maintaining ML models' practical performance in a dynamic environment. 
Note that continual learning is distinct from online incremental learning. The latter enables adaptation to new data as it arrives and does not include any mechanism to retain previous knowledge, whereas the former also considered retention of previously learned knowledge.
Notably, the importance of continual learning is underscored by its applicability across various domains, ranging from computer vision and natural language processing to robotics and autonomous systems~\cite{CL_ComputerVision, CL_robotics}. For instance, in computer vision, continual learning is crucial for developing systems capable of adapting to changing visual contexts and learning new object categories without forgetting previously learned ones. Similarly, in the field of robotics, continual learning enables robots to acquire new skills and adapt to diverse environments over their operational lifespan.

Despite the great potentials of continual learning, its applications in particle accelerators has not been fully explored and there still lacks a dedicated avenue for exchanging the underlying ideas. 
To fill this void, this paper seeks to present a review of the challenges and opportunities in continual learning as it applies to particle accelerators. 
We critically evaluate existing continual learning methods, assessing their relevance and applicability within the particle accelerator domain. 
We describe potential future avenues for research in continual learning, providing concrete applications that can propel advancements for ML in particle accelerators.
Our analysis aims to identify key challenges and opportunities in application of continual learning in accelerators, thereby paving the way for future advancements and applications in this domain. 
This paper mainly focuses on applications of ML for anomaly prediction/detection, optimization, real-time control, surrogate models and digital twins, and virtual diagnostics to provide valuable insights to adapt ML models applied in these use-cases with continuously drifting accelerator data.

The remainder of the paper is organized as follows.
Section~\ref{ch:background} provides a brief background on existing applications of ML in particle accelerators, and their limitations,
Section~\ref{ch:deploymentChallenges} describes challenges associated with deployment of ML solutions in particle accelerators,
Section~\ref{ch:continuallearning} presents a short but comprehensive review on continual learning and the promises it holds. 
Section~\ref{ch:deployableML} presents potential areas within particle accelerators where continual learning methods can help mitigate the deployment challenges associated with data drift. We also highlight the feasibility of different continual learning methods for various applications given constraints linked to individual applications.
We close the discussion with a conclusion and future outlook in section~\ref{ch:conclusion}.


\section{AI/ML in Particle Accelerators}
\label{ch:background}

Many areas within particle accelerators have shown remarkable improvement in both performance and time-efficiency with assistance from ML,  including anomaly detection and prognostication to avoid downtime and equipment damages, optimization of design, optimization of operation and real time control, beam tuning and shaping etc.
However, the potential of these solutions has not been fully realized in operational usage due to challenges associated with data drifts, ML lifecycle management and time sensitivity.
In what follows, we first review some of the recent developments at the intersection of ML and particle accelerators, and describe how they fit in the context of continual learning.

\subsection{Anomaly Detection}

\textbf{Auto-encoders:} 
Particle accelerators produce a large volume of data recorded by various diagnostics and monitoring sensors. 
A majority of this data is from normal operation and a small amount of data represent anomalies. 
Normal machine behavior can also be mimicked by simulations. This leads to a natural class imbalance for anomaly detection applications.
Auto-encoders (AE)~\cite{10.5555/104279.104293} or Variational Auto-encoders (VAE)~\cite{https://doi.org/10.48550/arxiv.1312.6114} are unsupervised (or semi-supervised in some cases) ML algorithms that can be trained on imbalanced datasets without labels, by leveraging large amount of existing normal data  and  clustering the latent space to detect anomalies. 
The underlying assumption is that the similarity between normal samples would lead their latent representations to be close together, whereas latent representation of anomaly samples would be far away from the normal cluster(s).
In this direction, a resilient VAE model has been developed to detect anomalies at Linac Coherent Light Source (LCLS) at SLAC National Laboratory (SLAC) and shown to perform well when trained on a large volume of normal data and a small set of anomaly data~\cite{VAE_Slac}.
On the other hand, AE/VAEs can also be trained only on normal data (semi-supervised) to capture the features that build the normal data samples and reconstruct them with a very low error. During inference, it is expected that the model would have high reconstruction error on anomaly samples as they are expected to be different from the training data. 
Here, instead of leveraging latent representation, reconstruction error is used as a classification indicator to detect anomalies.
One such model was developed at Brookhaven National Lab (BNL) to detect anomalies in air conditioning system of their accelerator~\cite{gao:ipac2023-thpl013}. 

Particle accelerators are actively tuned and acquire multiple different configuration settings during normal operation. 
They also have multiple sub-systems producing similar data, where the normal data collected from different sub-systems or different run conditions can vary producing multiple data distributions.
In such cases, conditional VAE is leveraged at the Spallation Neutron Source (SNS) accelerator for errant beam prediction in Super Conducting Linac (SCL)~\cite{Rajput_2024}. 
In a similar effort, conditional VAE is leveraged to learn the distribution of the data originating from different 
High Voltage Converter Modulators (HVCM) to detect anomalies~\cite{ALANAZI2023100484}.
Additionally, a study at SNS accelerator has shown that contrastive learning performs much better at anomaly prediction than reconstruction based methods such as AE~\cite{Blokland:2021onk}.
These applications has beed demonstrated on using curated offline dataset or during short online studies where conditions are relatively constant. 
Deploying these solutions for longer term online applications would requires continual model adaptation as AE/VAE can lead to higher reconstruction or misleading clustering on drifted data.

\noindent
\textbf{Contrastive Learning:} 
Contrastive learning has been mainly developed for few shot learning in ML~\cite{yang2022fewshotclassificationcontrastivelearning} where the feature extraction knowledge can be transferred from one dataset to another before fine-tuning with a minimal number of samples (few shots) from the new dataset.
One intrinsic property of contrastive learning methods, such as Siamese Neural Network (SNN)~\cite{chicco_siamese_2021}, is that it can also address class imbalance in the training data. Specifically, 
SNN requires a pair of inputs and learns to predict similarities between the two input samples.
By exploiting this property, a large amount of normal data can be paired with a small number of anomaly samples to produce the positive class, and a roughly equal number of pairs can be created with normal-normal data as negative class samples.
Recent studies at SNS accelerator have shown that SNN can outperform semi-supervised AE/VAE~\cite{Blokland:2021onk, Rajput_2024}. 
The study was further improved by developing conditional SNN which improve the predictions of errant beams at SNS when considering multiple beam setting parameters.
However, the study also identified a need for continual learning to maintain model performance on time varying data drifts~\cite{Rajput_2024}.

\noindent
\textbf{Other} techniques for anomaly detection on accelerators include supervised classification based feed-forward neural network, explored in pressure systems at SuperKEKB accelerator~\cite{PhysRevAccelBeams.27.063201}. 
For time-series data, Recurrent Neural Network (RNN)~\cite{schmidt2019recurrentneuralnetworksrnns} models have been developed to detect anomalies in Radio Frequency (RF) cavities at Continuous Electron Beam Accelerator Facility (CEBAF)~\cite{Solopova:IPAC2019-TUXXPLM2}.
Another effort at SNS studied detection of anomalies in HVCM system using Long Short Term Memory (LSTM) networks~\cite{pappas:ipac2021-thpab252}.
These applications use neural network that are shown to have performance degradation on drifting data~\cite{Concept_drift}, as such continual learning can enable their long term usage.


\subsection{Optimization and Control}
Particle accelerators require active tuning of various equipment before and during operation. 
Human operators are often tuning these machines for hours to achieve required beam delivery.
ML based optimization and control algorithms has been explored to help alleviate these challenges ~\cite{duris_bayesian_2020,roussel2021turnkey,LeemannPRL2019,pmlr-v162-kaiser22a}.
Various algorithms may incorporate models that persist across multiple tuning sessions, thereby presenting a topic of interest from the perspective of continual learning. 
For example, with standard Bayesian optimization (BO), a new model is learned from scratch during each optimization run. 
With advance enhancements BO can also leverage priors from learned system models, which may need to be updated over time.
This is similar to Model Predictive Control algorithm that often uses a separately-developed system model that persists along with a policy model. 

One application of model updating in online control for accelerators is presented in \cite{ALS_Finetuning_Approach}, where a model-based approach is employed to implement feed-forward correction for regulating beam sizes in the presence of varying insertion devices. 
In this framework, a base model is trained on a diverse set of data distributions. 
When encountering a new data distribution, the same base model undergoes fine-tuning to adapt to the updated conditions. 
This methodology aligns with principles of online learning. 
However, it does not address the issue of catastrophic forgetting~\cite{CF}, as the specific application does not necessitate retaining prior knowledge. 
Instead, the model is only required to reflect the most current state of the machine.
\\
\\
\noindent
\textbf{Bayesian Optimization (BO) }
BO~\cite{GP_bible} is a powerful framework designed to optimize black-box models, particularly in the context of system parameters such as magnet settings in accelerators.
The algorithm intends to find optimal settings by effectively mapping input variables to a target objective function.
Utilizing Gaussian Processes (GPs)~\cite{GP_bible}, BO not only estimates the function values but also quantifies uncertainties associated with predictions.
One of the significant advantages of BO is its ability to operate with limited data, enabling the development of a local system model from the ground up. 
This local model enhances the efficiency of the optimization process, allowing BO to learn from fewer samples compared to traditional methods.
Additional information about uncontrolled inputs can be incorporated into ``contextual'' BO, but this is limited by the computational scaling of GPs. 
BO also enables learned output constraints to easily be incorporated into tuning, which is appealing for deployment on unexplored systems or setups \cite{roussel2021turnkey}. 
BO has been used to optimize free-electron laser at SLAC~\cite{duris_bayesian_2020, McIntireIPAC16}.
In addition, BO has been explored for optimization tasks such as beam injection process~\cite{PhysRevAccelBeams.26.034601, Awal_2023}, for recoil mass separator~\cite{PhysRevAccelBeams.25.044601}, and accelerator tuning~\cite{MORITA2023168730}.

Fast and accurate system models can be used to provide prior information, correlated and learned kernels for BO to improve learning and increase convergence speed during online tuning~\cite{boltz2024more, hanuka_physics_2021, DKL}.
Continual learning could be highly relevant for keeping system models used as priors, physics-informed kernels, and similar learning boosts up to date with data drifts.

\noindent
\textbf{Reinforcement Learning}
Deep Reinforcement Learning (DRL) has emerged as a powerful approach in various control applications, particularly in robotics~\cite{RL_robotics_survey}. 
Its inherent alignment with time- and sequence- dependent control problems makes it well suited for identifying optimal control actions while considering the future evolution of the system. 
Although DRL typically requires more development overhead, it offers the advantage of continuous control capabilities. 
Once trained, DRL can provide solutions more rapidly than BO, and can also ingest large amounts of information from previous machine interactions or archived data.
This allows DRL to leverage previously-learned correlations across different machine operating conditions, enhancing control performance. 

DRL has shown promising results for both optimization and control applications within particle accelerators.
Notable applications include its use for optimizing heat load and trip rates at Jefferson Lab CEBAF~\cite{rajput2023multi}, and for booster control at Fermilab~\cite{PhysRevAccelBeams.24.104601}.
Researchers have illustrated the potential of RL algorithms to enhance the efficiency of particle accelerator operation, with optimization of beam injection at COSY facility in Germany~\cite{awal2024injectionoptimizationparticleaccelerators}, optimization and control at FERMI FEL facility in Italy~\cite{RL_FERMI_FEL}, proof of concept autonomous sub-system control~\cite{pang2020autonomouscontrolparticleaccelerator}, and beam line control~\cite{Schenk_2024} at CERN.
While the use of Multi-Objective RL (MORL) has been limited in the particle accelerator domain, it can naturally provide dynamic control capabilities. 
MORL has been evaluated for simultaneous optimization of heat load and trip rates in CEBAF linacs at Jefferson Lab~\cite{MOO11_Kishan}.
A recent study on a multi-objective constrained optimization problem concluded that differentiable RL can outperform traditional MORL on large dimensional problems~\cite{rajput2024DDRL}.

One of the key limitations of traditional DRL approaches in these applications is their inability to adapt efficiently to evolving machine conditions without extensive retraining or catastrophic forgetting.
This is also evident from the fact that most of the above discussed DRL based solutions are yet to be seen as a long term deployed solution.
Traditional DRL typically relies on continuous training with experience replay, where past experiences are stored and reused to stabilize learning. 
However, this approach is limited in dynamic environments, as it does not inherently adapt to evolving conditions and can suffer from catastrophic forgetting when faced with new operational scenarios. 
In contrast, continual learning enables DRL controllers to retain past knowledge while integrating new information allowing for real-time adaptation to drifting data. 

\noindent
\textbf{Model Predictive Control}
Similar to DRL, Deep Model Predictive Control (DMPC) is also suitable for time- and sequence- dependent control problems, where the goal is to find optimal control actions by considering the future evolution of the system. 
To achieve this, DMPC typically requires a sufficiently accurate and fast-executing system model.
This approach enables additional constraining of control behavior and provides more directly interpretable actions. 
Moreover, DMPC has been successfully applied to various applications in particle accelerator space where system model(s) are available. 
Some of the notable applications include power supply control~\cite{DMPC_powerSupply}, data driven DMPC to refine the multi-turn injection process into the heavy ion-synchrotron SIS18 using Gaussian process model~\cite{DMPC_SIS18}, application of DMPC on inner triplet heat exchanger unit of the large hadron collider at CERN based on a nonlinear model derived from physical relationships~\cite{DMPC_LHC}, ion-source control for beam tuning at RFT-30 cyclotron~\cite{DMPC_ionSource}, and Radio Frequency Quadrupole resonant frequency control in Proton Improvement Plan II at Fermi lab~\cite{Edelen_RFQ}.
The type of system model used in DMPC can vary between empirically tuned physics models, to learned neural network models, to Gaussian processes. 

Continual learning can enhance the adaptability of neural network policy models with changing conditions.
In addition, it is also applicable where neural networks are used as a system model and can provide great benefits to keep it up-to-date with drifting data.

\subsection{Surrogate Models and Digital Twin}
\begin{figure}
    \centering
    \includegraphics[width=0.85\linewidth]{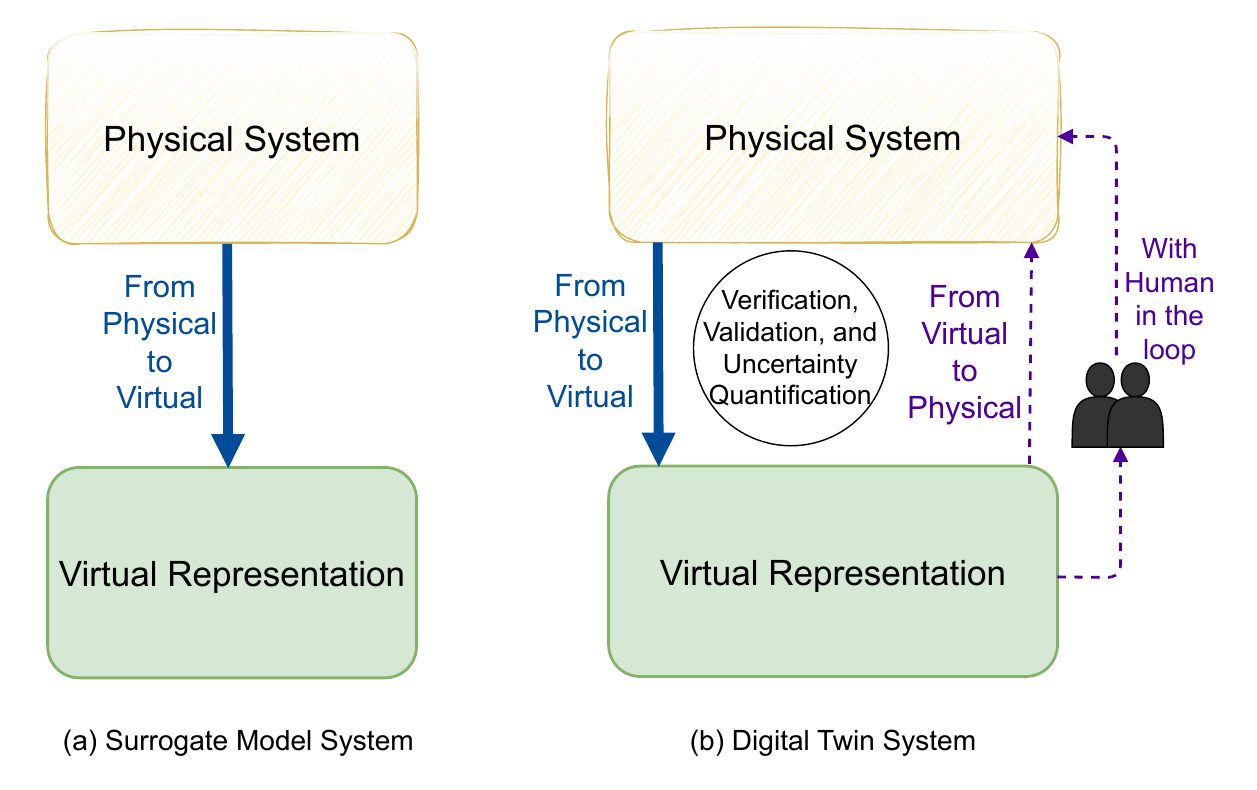}
    \caption{Differences between a typical surrogate model and digital twin. (a) Schematic Diagram of surrogate model system. (b) Schematic diagram of a digital twin system with two way information flow with a possibility of having human-in-the-loop.}
    \label{fig:Surrogate-vs-DT}
\end{figure}

A surrogate model is a computational model that is used to approximate the behavior of a more complex, detailed, and often computationally expensive process.
Surrogate models are often used to replace physical processes that are resource-intensive, time consuming and/or expensive. 
On the other hand, a Digital Twin (DT)~\cite{DigitalTwinRef} provides a real-time, dynamic representation of a physical entity, process, or system. 
It integrates real-time data with simulation models to monitor, predict, and optimize the performance of the physical counterpart.
As demonstrated in Figure~\ref{fig:Surrogate-vs-DT}, surrogate models often have a one-way information flow where a data driven model is trained on measurements obtained from the physical machine. 
In contrast, a DT has a two-way information flow, where not only the virtual model is continuously adapted to mimic real-time physical system behavior but it is also used for decision making to improve the physical system. This process often may involve human operators in the loop.

In particle accelerators, both data and physics driven models have been developed and used for various purposes such as training of a control algorithm, diagnostics and pre-emptive maintenance. 
Simulating particle accelerator processes in digital space is both computationally expensive and time consuming. There are a number of physics based beam dynamics simulations~\cite{GPT, virtual_accelerator, Bmad2006, pyOrbit}, however, discussion of purely physics driven simulations is out of the scope of this paper.

To reduce computational resources requirements and execution time, many particle accelerator processes are being represented as data driven surrogate models~\cite{UQ_Surrogate_Andy}.
These models have mainly been used to train control algorithms which are infeasible to train on the real physical machine due to various constraints.
For example, a robust digital twin was developed at Fermi National Accelerator Laboratory (FNAL) to train a RL agent~\cite{DT_RL_Schram}.
A surrogate model is also used at ISIS Neutron and Muon Source to assist in tuning of beam losses~\cite{surrogate3_isis}.
Neural network based surrogate models of RF quadrupole ~\cite{surrogate4}, and an encoder-decoder neural network based surrogate model of Medium Energy Beam Transport (MEBT)~\cite{surrogate2} have been shown to be effective in simulations.
Uncertainty aware surrogate models have been studied at FNAL for their booster complex and shown good performance on detecting out of distribution samples~\cite{Surrogate_UQ_kishan}.

Similarly, DT can be used to test accelerator physics programs before applying in the field. 
Several DTs have been developed in particle accelerator space~\cite{10752039, edelenprogress, isolan2022digital}.
Developing reliable continual learning approaches for digital twins in accelerators is a major need, especially for systems which experience substantial drift over time or frequently change experimental setups.

\subsection{Virtual Diagnostics}

Virtual diagnostics provide predictions of machine behavior that cannot be directly measured (at all or continuously). 
For example, the X-band Transverse deflecting mode CAVity (XTCAV) at FACET-II, and beam viewers at Upgraded Injector Test Facility (UITF) at Jefferson lab are destructive to the beam, and thus beam cannot be delivered to users and monitored on these sensors simultaneously.  
Virtual diagnostics can use other measured data to predict the output of the unavailable diagnostic.

Virtual diagnostics can use supervised learning to map measured inputs to predicted outputs (e.g. see~\cite{emma2018machine, SanchezGonzalez2016AccuratePO, DESY_VirtualDiagnostics_Simulation}). 
Other approaches adjust a learned system model that is adjusted based on observations to then provide a quasi-real-time estimate of other beam properties which are either inaccessible or require destructive measurements (e.g. see \cite{scheinker}). 
Distribution shift over time is a major challenge for practical application of virtual diagnostics in practice. 
Users must have assurance that the predictions are going to continue to be accurate for use in analysis and control. 
\\
\\
Evident from the above discussion, with development of many ML models for particle accelerators, researchers have shown that ML holds tremendous potential to enable autonomous operation of particle accelerators. 
However, it is evident that continual learning is critical to enable long term deployment of ML in particle accelerators and there are still several challenges associated with it as described next.

\section{Deployment Challenges in the Context of Continual Learning}
\label{ch:deploymentChallenges}

Despite significant advancements in AI and ML for particle accelerators, the number of implemented approaches remains notably behind the apparent progress. 
This gap arises from both algorithmic and practical challenges: some cases require fundamental improvements to AI/ML techniques to ensure reliable deployment, while others involve mature techniques that still need the appropriate software infrastructure for effective implementation and integration into operations. 

A significant technical challenge is data distribution shifts due to changes in the equipment settings, drift in the data due to unmeasured/unknown parameters such as machine degradation or replacement of equipment, and changes in the environmental conditions. 
The presence of distribution shift means an ML solution cannot simply be created once and deployed; it needs to be monitored and updated over time.
Addressing data distribution shift effectively requires both algorithmic improvements and infrastructure for continual learning,


\subsection{Data Drifts and Distribution Shifts}
In complex machines, such as particle accelerators, any deviation in behavior of one or more pieces of equipment can quickly propagate and shift the distribution of the measurements downstream.
Variations in initial conditions, changes in settings for optimization, and environmental factors can lead to shifts in the data that ML models were initially trained on. 
These shifts can adversely affect model performance, resulting in decreased accuracy and reliability in predictions.
This is due to concept drift as the relationship between input features and target variables changes over time.
As such, if the data distribution shifts, the machine learning model may extrapolate incorrectly or may not represent the up-to-date concept, leading to reduced accuracy and performance. 
Detecting and addressing concept drift is crucial to maintaining the reliability of ML outputs. 
Continuous monitoring, along with adaptive algorithms and retraining strategies, is essential to identify and mitigate the effects of data- and concept- drift, ensuring that ML models remain robust and effective in interpreting evolving experimental data.
Data distribution shifts in particle accelerators can broadly be classified into the following two categories.


\subsubsection{\textbf{Data distribution shifts due to measured parameters}}
\label{sec:drift_measuredparameters}
When the beam parameters and/or equipment settings are changed, they shift the measurements.
These types of distribution shifts are usually easier to detect and even predict in some cases as the changes in beam setting knobs can be quantitatively measured.
However, these changes can still make an ML model obsolete if the shifts are significant.
One of the ways to minimize such effects is to use conditional ML models where the beam settings and other relevant parameters are used as conditional inputs to the model(s)~\cite{Rajput_2024}.
This helps the model learn the correlation between the data samples from different settings and effectively interpolate to new settings. 
However, if the settings go out of distribution with significant shift from the training data distribution, the model can still fail and requires to be fine-tuned.

\begin{figure}
    \centering \includegraphics[width=0.99\linewidth]{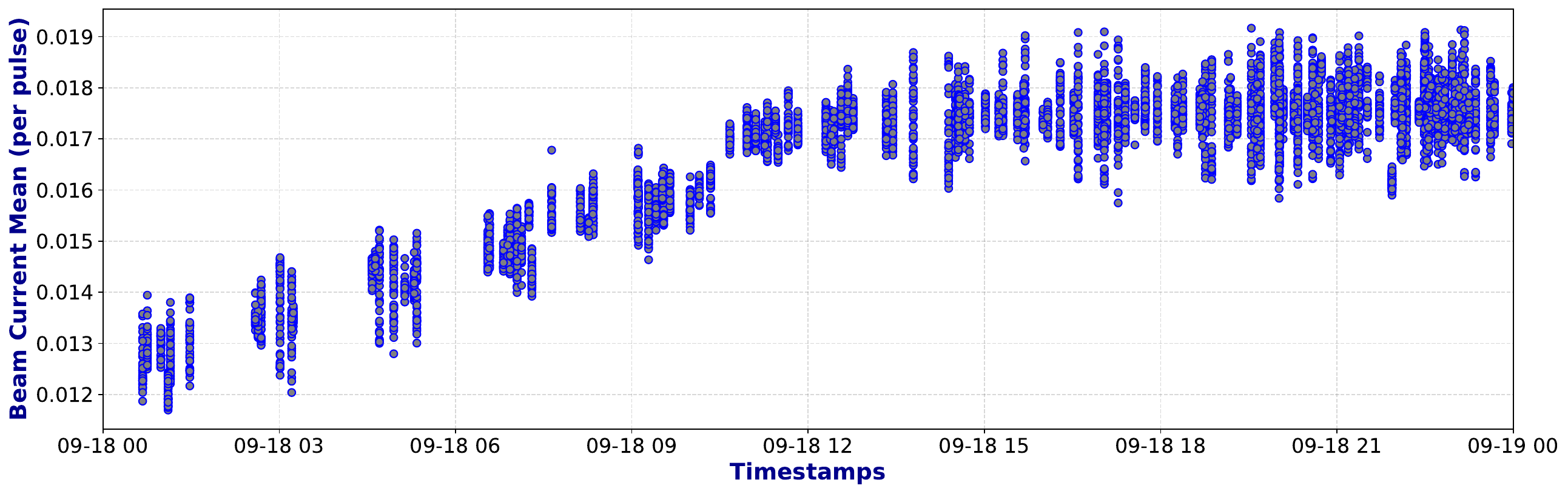}
    \caption{Drift observed in beam current data obtained at SNS accelerator. The y-axis represent mean of the beam current waveform per macro-pulse. Though, it is a harsh summarization of full macropulse about 100K measurements long, it clearly demonstrates the drift in the measurements.}
    \label{fig:drift}
\end{figure}

\begin{figure}
    \centering \includegraphics[width=0.99\linewidth]{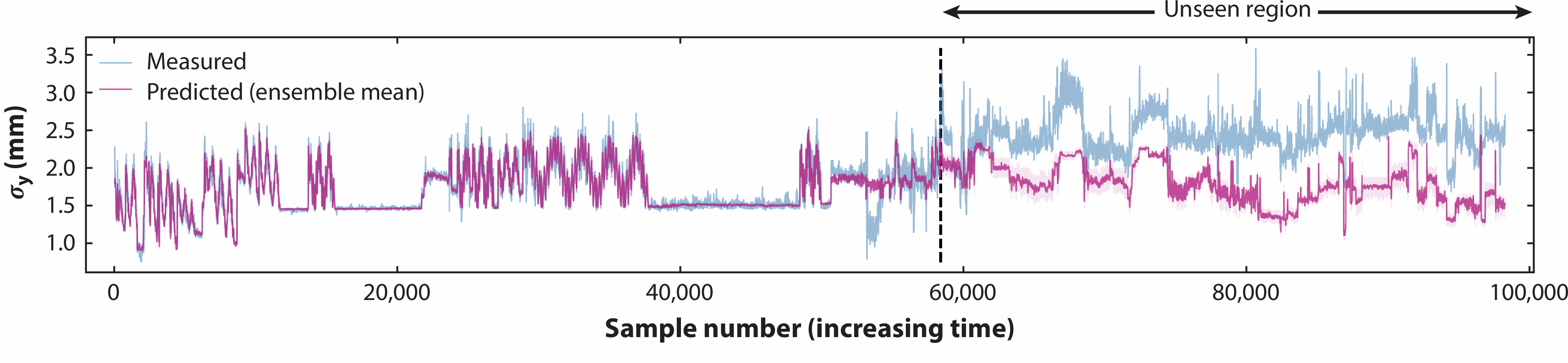}
    \caption{Impact on ML model prediction performance of drift due to slow variation in upstream inputs (laser distribution, rf cavity phase and amplitude) and potentially other unknown changes from a configuration change at a small accelerator. All data shown is test data, but during the period before the marker, there is corresponding training data. This shows the impact of drift on ML model predictions (in this case of the beam size). Reproduced from \cite{Edelen-MLforDesignAndControl}.}
    \label{fig:drift2}
\end{figure}

\subsubsection{\textbf{Drifts due to non-measured parameters}}
\label{ch:drift_nonmeasuredparameters}
Frequently particle accelerator data drifts due to unknown causes or if known relevant information is not recorded or non-quantifiable.
For example, particle accelerator data distribution typically shifts after a long shut down or a maintenance period due to changes in the equipment. 
Slow (and some rapid) drifts are still observed during continuous operation of particle accelerators. 
This type of drifts can arise due to equipment degradation, drifts in ion source output as plasma changes (depleting source), variation in the performance of klystron over time, high voltage system capacitors degradation, heating up of beamline components, other ambient temperature effects, etc.

These drifts are often substantial enough that operators (or automated algorithm) need to re-tune the accelerator. 
Further, the sources of drift may not have a direct measurement available or may be challenging to determine out of the many signals. 
For example, at SPEAR3 accelerator, the orbit settings needed to often be re-tuned due to drift; an investigation of sources of drift found the closest correlation with an ambient temperature sensor, indicating bulk thermal behavior and expansion and contraction of the injector and ring may be responsible for the orbit drift \cite{SPEAR3_Drift_Temp}.

Figure~\ref{fig:drift} shows drifts due to both known beam setting changes and gradual drifts caused by unknown factors in the measured beam current data at SNS accelerator. 
The figure summarizes a time series of beam current data for each pulse into its mean for an easier visualization.
Similarly, Figure~\ref{fig:drift2} shows impact of gradually drifting data due to slow variation in upstream inputs (laser distribution, RF cavity phase and amplitude) and potentially other unknown changes on ML model predictive performance. 
It clearly demonstrate that the ML model fails when the data distribution drifts away from the training data distribution.

\subsection{Low Latency Requirements, Large Volume of Data, and Large Number of Variables}
Particle accelerators are often running at high repetition rates, for example a pulsed accelerator such as SNS runs at ~60Hz, and CEBAF provide a continuous beam of electrons making one round of $\frac{7}{8}$ mile long path in less than a microsecond.
Deployable ML models should provide timely responses in order to be actionable~\cite{Rajput_2024}.
ML models deployed for decision making require precise synchronization with accelerator operations.

On the other hand, accelerators produce huge amounts of data collected from many diagnostics sensors such as Beam Position Monitors (BPMs)~\cite{BPM}, and Differential Current Monitors (DCM)~\cite{Blokland:2021onk}.
Anomaly prediction and just-in-time control models require swift processing of the data.
Collection, tagging, synchronization across sensors and storage of large volume of data requires a significant amount of resources including software infrastructure and storage server.
Achieving low-latency processing can be difficult with the computational demands of large deep learning models.

Particle accelerators are paradoxically limited in number of detailed diagnostic sensors due to many reasons including cost, physical space, etc. 
However, accelerators also in some cases store hundreds to thousands of measurements at each moment in time, spanning devices across the accelerator. 
The high dimensionality of these systems can make it challenging to build models that account for the numerous sources of potential changes over time. 
Data cleaning to prepare for model training and inference can also be challenging due to the dimensionality and the number of different ways signals on the accelerator can fail.
Many diagnostics devices operate in noisy environments, leading to significant signal interference.
These factors complicate data curation and the training of ML models. 
A key challenge is balancing the use of more available signals to improve model performance against the risk of introducing failure points if those signals become unreliable.
Additionally, distribution shift over time can make it difficult to maintain a consistent model unless the problem is explicitly treating as time-dependent.
These factors play a critical role in selection of appropriate continual learning method as described further in this paper.

\subsection{Infrastructure to Support Continual Learning and Deployment}
While AI/ML research is often well-funded, the software and infrastructure development required to robustly deploy AI/ML in operations is often neglected. 
For some applications, the accelerator computing, hardware, and networking infrastructure must be enhanced to enable the data tagging, storage, and movement capabilities to support AI/ML. For routine deployment into operations, a workflow to update AI/ML models must be implemented in such a way that it can be used by physicists and operators and not just by experts. Integration of compute-intensive AI/ML approaches into operation can require linking the accelerator control system to High Performance Computing (HPC) systems. 

Existing particle accelerators and their infrastructure were not designed to accommodate ML deployment .
New infrastructure will be required to synchronize data from multiple sensors for any global ML model deployment.
Furthermore, many particle accelerators rely on established software systems that may not be designed to accommodate modern ML frameworks. 
Integrating ML models into these legacy systems can require extensive modifications, which can be challenging and time-consuming. 

\section{Continual Learning}
\label{ch:continuallearning}

Continual learning is a dynamic process of continuously updating and refining models to adapt to changing conditions, thereby mitigating the impact of concept drift~\cite{Thrun1998, PARISI201954}.
Continual learning is an active area of research within ML, presenting a paradigm shift from traditional static learning approaches. 
In contrast to traditional machine learning paradigms, where models are trained on static datasets and assumed to function within stable environments, continual learning requires that models possess the capability to seamlessly adapt and acquire knowledge from a continuous data stream over an extended time-frame, all while preserving insights gained from previous experiences. This ability replicates the mechanisms of human learning, allowing machines to progressively accumulate knowledge, adjust to evolving tasks, and maintain performance within dynamic environments.
This iterative approach can be linked to a maintenance regimen for ML models in production environments, where the models are continuously updated to maintain optimal performance.
As ML models are updated with new data from a different distribution than the previous training data, it introduces a stability vs. plasticity dilemma~\cite{stability-plasticity}. 
Here, stability refers to the ability of the model to retain its performance on previous data distributions, while plasticity refers to the flexibility of the model in learning new data or concept. 
An ideal algorithm would maximize both, however achieving a good balance between stability and plasticity is challenging. 
When given new information, a model with high plasticity may forget previously acquired knowledge. 
This phenomenon is referred to as catastrophic forgetting~\cite{CF}.
On the other hand, if a model has high stability, it may be too rigid and may not be able to acquire new knowledge. 


\subsection{Data and Concept Drifts}
 
Data drift refers to changes in the statistical properties of input data, meaning that the features or distributions of the input data differ from those encountered during ML model training~\cite{concept_drift1}. 
This can arise due to various factors such as seasonality, changes in user behavior, or external events~\cite{concept_drift2}. 
Concept drift, on the other hand, occurs when the underlying relationship between the input features and the target variable evolves~\cite{learning_in_concept_drift}. 
For instance, a model trained to predict customer preferences may become less accurate as consumer trends shift. 
Both types of drift necessitate continuous monitoring and updating of machine learning models to maintain their accuracy and reliability, prompting the development of adaptive techniques such as retraining, ensemble methods, and drift detection algorithms~\cite{learning_in_concept_drift, learning_in_concept_drift2}. 
Addressing these issues is crucial for ensuring that models remain effective and relevant in dynamic environments.

Data drifts can be categorized based on their periodicity, speed, and scope. 
\textbf{Periodic} drifts exhibit regular recurrence, while \textbf{irregular} drifts lack any discernible pattern. 
Based on speed, drifts can be classified as sudden, gradual, or incremental. 
\textbf{Sudden} drifts occur rapidly, leading to a new data distribution. 
\textbf{Gradual} drifts involve a transitional phase where the original and new distributions coexist. \textbf{Incremental} drifts occur slowly and subtly, making them challenging to detect. 
Based on scope, drifts can be classified as local or global. 
\textbf{Local} drifts impact specific subsets of the data, while \textbf{global} drifts affect the entire dataset. 
Local drifts often indicate changes in the data collection or shifts in the underlying sub-processes producing data for the local shifted region. 
Whereas global drifts implies a fundamental shift in the overall characteristics of the data, impacting all regions or subsets equally. 
\textbf{Recurrent} drifts follow repeating patterns, often driven by seasonal variations or cyclical trends. 
These types of drifts are often correlated with Periodic drifts. 
\textbf{Evolving} drifts are continuous and non-periodic, reflecting gradual changes in the underlying data-generating processes. 
Evolving drift can result from various factors such as changing user preferences, shifts in market dynamics, or gradual environmental changes.


Detecting drift in machine learning is essential for maintaining the performance and reliability of models deployed in dynamic environments where the underlying data distribution may change over time. 
Below we briefly introduce approaches proposed in the literature to detect the drifts.

\subsubsection{Window-based methods}
\hfill
\\
\noindent
A number of window based drift detection methods are explored by the researchers. 
Simple methods include sliding window statistical tests such as Kolmogorov-Smirnov (KS) test~\cite{KS_test}, Chi-square test~\cite{chi_square}, Wasserstein distance~\cite{wasserstein_distance} that can be used to detect any deflection from normal trend with a threshold.
In a slightly different approach sliding window statistical distance (Jensen-Shannon Divergence~\cite{JSD}, Kullback-Leibler (KL) Divergence~\cite{KLD}, Euclidean distance, earth movers distance etc.) can be compared with a fixed reference window.
ADWIN (Adaptive Windowing)~\cite{window_drift_detection} is another drift detection method used in data streams to monitor changes in the underlying statistical properties of the data. 
It is particularly useful for detecting concept drift, which refers to changes in the relationship between input features and target variables over time.
The basic idea behind ADWIN is to maintain a sliding window of fixed size over the incoming data stream and monitor statistics within that window. 
As new data points arrive, ADWIN continuously updates its assessment of whether the data distribution within the window has remained stable or if a significant change, indicative of drift, has occurred.
ADWIN is known for its ability to detect drift with low computational overhead and adapt to varying rates of change in the data stream. 
It has been widely used in applications such as online learning, real-time monitoring, and adaptive systems where timely detection of concept drift is critical for maintaining model performance.

\subsubsection{Predictive Error-based methods}
\begin{figure}
    \centering
    \includegraphics[width=0.85\linewidth]{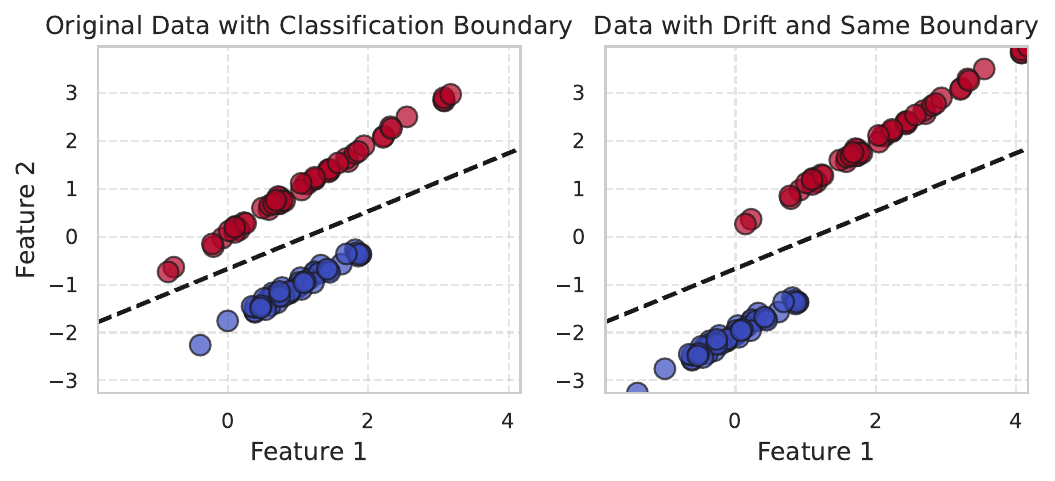}
    \caption{Data drift within same classes (no concept drift)}
    \label{fig:data-drift-dummy}
\end{figure}

\hfill
\\
\noindent
It is important to note that the data drift do not always implies concept drift or model degradation. 
Figure~\ref{fig:data-drift-dummy} shows one such example where data drifts within the individual classes. In this example, data drifts, however no concept drift is observed as the previous classifier is still valid. 
Drift detection methods that only focuses on data distribution may trigger unnecessary model re-training in such cases.
This is the reason why predictive error based drift detection methods has become more popular. 
Predictive error based methods can also detect other potential issues with the data/inference pipelines that may cause performance degradation.
As such, predictive error based drift detection methods are proposed.
One such method is called Drift Detection Method (DDM)~\cite{error-based-drift-detection} which is later extended to Early Drift Detection Method (EDDM)~\cite{EDDM} and McDiarmid Drift Detection Method (MDDM)~\cite{MDDM} to improve the performance on gradually drifting data. 
In addition, window based methods discussed above can also be used to monitor the predictive error-rate with a sliding window and trigger when the rate crosses a pre-defined threshold.

\subsubsection{OOD  uncertainty based drift detection}
\hfill
\\
\noindent
An OOD uncertainty-based data drift detection method operates by assessing the uncertainty associated with model predictions on incoming data samples. 
The core idea is to analyze whether the incoming data points belong to the same distribution as the training data, using uncertainty metrics derived from the model's predictions.
ML models has shown to provide OOD uncertainties with Bayesian Neural Networks~\cite{Goan2020}, Deep Gaussian Process Approximation~\cite{liu_simple_2020, schram2023uncertainty}, and Gaussian Processes~\cite{GP_UQ}.
In this approach, the uncertainty scores produced by the model are continuously evaluated for the predictions on new data points. 
If the uncertainty exceeds a predefined threshold, it indicates that the data point may be OOD, suggesting a potential shift in data distribution. 
This threshold can be calibrated based on historical performance metrics or statistical analysis of uncertainty scores~\cite{OOD1}.
OOD based uncertainties are not fully leveraged yet, however it has the potential to be very useful in validating predictions of the models, especially in real-time decision making processes.

\begin{figure}[h]
    \centering
    \includegraphics[width=\textwidth]{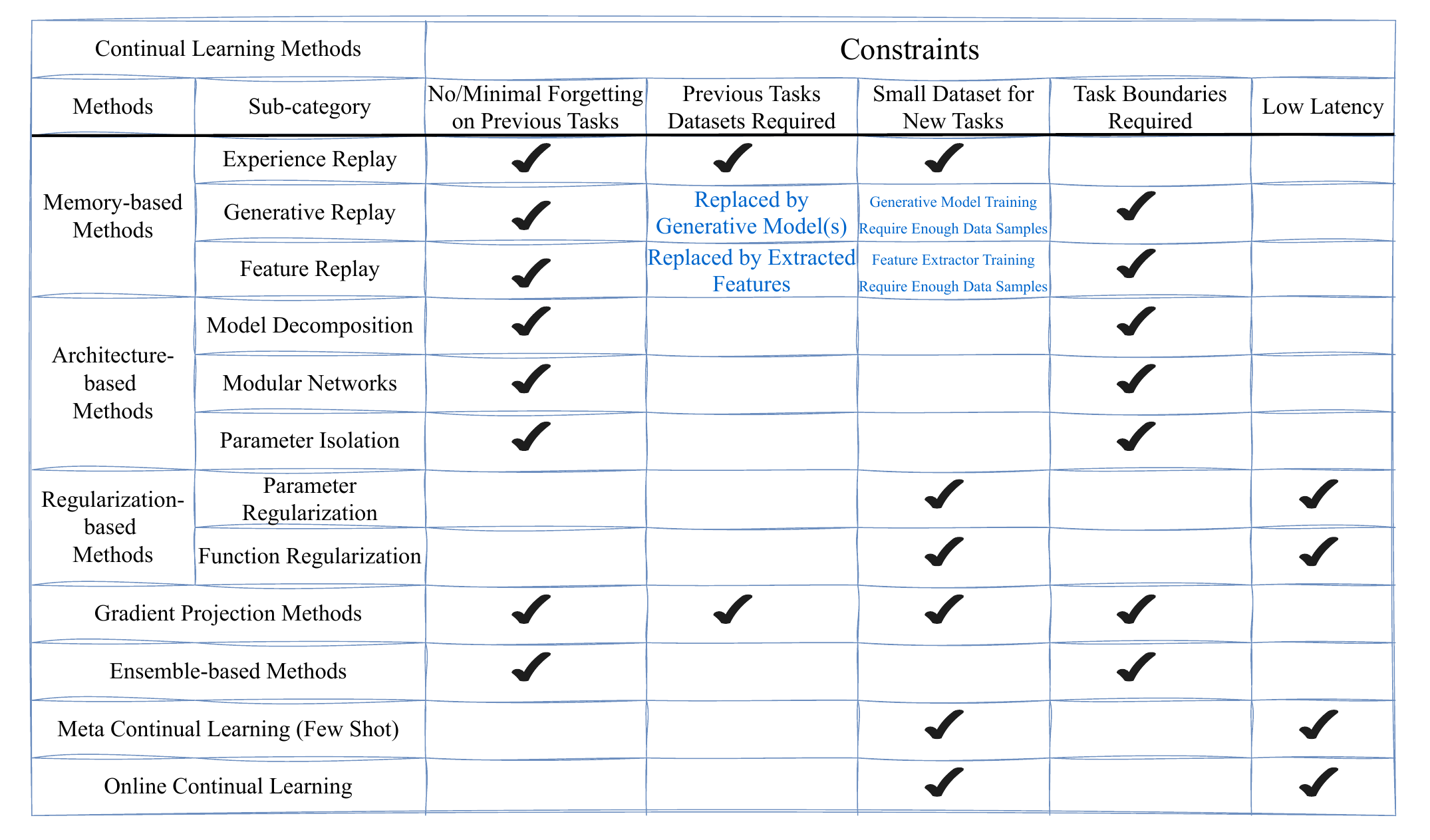}
    \caption{A guide to selection of appropriate continual learning method(s) given system constraints}
    \label{fig:methods_table}
\end{figure}

\subsection{Types of Continual Learning Approaches}
The continual learning methods listed in Figure \ref{fig:methods_table} are described below with examples of their existing usage.

\subsubsection{Memory-based Approaches}
\hfill
\\
Memory-based continual learning methods have shown promising performance in addressing the challenges of catastrophic forgetting, where a model trained on new data may lose the knowledge acquired from previous tasks. 
These methods incorporate past experiences by replaying or revisiting old data during the training process, thus reinforcing previously learned representations while learning new information.

One common approach is \textbf{experience replay}, which involves storing a subset of past data and integrating it into the training of new tasks \cite{pmlr-v119-fedus20a, ER_for_CL}. 
This technique has been effectively utilized in reinforcement learning, where agents retain and reuse past experiences to improve performance \cite{mnih2015human}. 
In the context of supervised learning, replay can be implemented by maintaining a buffer of data points from previous tasks and sampling them alongside new task data during training.
Priority experience replay~\cite{PER} enhances learning efficiency by replaying experiences based on their importance, rather than sampling them uniformly. 
Samples that indicate greater learning potential, are prioritized, ensuring the model focuses on more informative samples.
Another approach that focuses on exploiting negative experiences to extract useful knowledge is hindsight experience replay~\cite{HER}, which has been explored in practical application of continual learning for sequential object manipulation~\cite{luo2022relayhindsightexperiencereplay}.

Due to high memory requirements of experience replay methods, \textbf{generative replay} approach has been developed.
These methods typically employ generative models, such as Variational Autoencoders (VAEs) or Generative Adversarial Networks (GANs), to approximate the data distribution of past tasks and replay samples from this distribution as if they were real~\cite{generative_replay_in_CL, generative_replay}. 
By doing so, the model can consolidate knowledge from previous tasks while learning new ones, enabling it to perform well across all tasks without requiring access to original data. 
Generative replay is particularly beneficial in privacy-sensitive or resource-constrained scenarios, where storing real data from earlier tasks may not be feasible~\cite{three_scenarios_for_CL}.

Another replay approach is \textbf{feature replay} that addresses catastrophic forgetting by replaying learned feature representations instead of raw data, making them computationally efficient and less prone to privacy concerns. 
In this approach, the model stores or generates intermediate feature representations of data from previous tasks, which are replayed during training on new tasks to preserve knowledge while incorporating new information. 
Unlike generative replay, feature replay avoids the complexity of generating raw data and focuses on maintaining compact and semantically meaningful representations~\cite{feature_replay}. 
This method is particularly effective in architectures with encoder-decoder structures or pre-trained feature extractors, enabling scalability and efficiency in sequential task learning.

Recent advancements also include \textbf{multi-task learning with replay}, where a model is trained on multiple tasks simultaneously while periodically revisiting past tasks. This approach helps in maintaining performance across all tasks by ensuring that the model does not drift significantly from the knowledge gained previously \cite{Kirkpatrick_2017}. 
The integration of memory-efficient techniques, such as selective replay, further enhances the efficiency of replay-based methods by focusing on the most informative samples from past tasks \cite{hayes2019memoryefficientexperiencereplay}.

Overall, memory-based continual learning methods represent a promising direction for developing systems capable of adapting to new information while preserving prior knowledge, making them essential in applications requiring lifelong learning.
However, they have a high requirement of memory and/or computing resources and may not be suitable for low-latency tasks.

\subsubsection{Architecture based approaches}
\hfill\\
Architecture-based methods modify the neural network architecture to support the storage and retrieval of knowledge over time. 
Architecture-based methods can be classified into three categories, namely model decomposition, modular networks, and parameter isolation.
These approaches allocate specific components of the network to different tasks, enabling efficient retention of learned knowledge while incorporating new information.

Model decomposition divides the network into shared and task-specific components. 
Shared components capture common knowledge across tasks, while task-specific components focus on task-specific features. 
By decoupling these aspects, the model minimizes interference between tasks.
For example, Progressive Neural Networks (PNNs) maintain a set of architectures for different tasks, allowing new networks to leverage previously learned features without overwriting them. This method efficiently transfers knowledge by freezing the weights of previous networks while allowing new layers to learn new tasks~\cite{PNN0}.
PNNs has been demonstrated in robotic systems, where robots learned to manipulate objects in different environments by building on previously learned skills~\cite{PNN0, PNNs}.

Neural Modular Networks (NMNs) consist of multiple subnetworks that can be activated based on the task. Each module learns a specific aspect of the data, which helps in isolating the learning process and improving performance on diverse tasks~\cite{NMN0}.
A study showed that using modular networks for NLP tasks allows models to efficiently learn new languages while retaining proficiency in previously learned languages~\cite{LLM_Modular}.

Parameter isolation methods protect subsets of the network's parameters from being overwritten when learning new tasks. These approaches often rely on masking, freezing, or dynamically allocating parameters for each task~\cite{parameter_isolation}.

In general, architecture-based methods has minimal forgetting on previous tasks by nature as they assign different parts of the network to different tasks and freeze previous knowledge. However, these methods require task boundaries to be effective.

\subsubsection{Regularization based approaches}
\hfill\\
Regularization-based continual learning methods aim to mitigate catastrophic forgetting by imposing constraints on the model's learning process. 
Regularization-based methods can be broadly divided into two categories, namely parameter regularization and function regularization.
Regularization-based methods are advantageous in that they do not require explicit storage of past data, making them memory efficient. 
They allow the model to maintain performance across multiple tasks while learning new information. 
However, these approaches can introduce additional computational overhead and may require careful tuning of regularization parameters to achieve optimal performance \cite{review}.

Parameter regularization typically adds regularization terms on the gradients to protect the weights associated with previously learned tasks while allowing adaptation to new tasks.
Elastic Weight Consolidatin (EWC) \cite{ewc} is one of the pioneering methods in this category. 
It uses a quadratic regularization term that penalizes changes to the important parameters identified during training on previous tasks. 
Specifically, EWC calculates the Fisher information matrix to determine the significance of each parameter, thereby penalizing updates to parameters that are crucial for past tasks. 
This method enables the model to learn new tasks while preserving knowledge of previously acquired ones.
Memory Aware Synapses (MAS) \cite{mas} builds on the idea of EWC by introducing a more adaptive approach to regularization. 
Instead of using a fixed Fisher information matrix, MAS continually updates a measure of importance for each weight based on the current task. 
This allows the model to dynamically adjust its sensitivity to changes in the parameters as new tasks are introduced, providing a more flexible means of managing catastrophic forgetting.
Lifelong Learning via Regularization (LLR) \cite{llr} further extends regularization-based approaches by utilizing a combination of EWC and task-specific regularizers. This method applies multiple regularization terms, allowing the model to better capture the complexities of various tasks while balancing the trade-off between stability and plasticity in learning.

In contrast, function regularization operates at the functional level, aiming to preserve the model's output behavior (the function it represents) for previous tasks. 
This is achieved by explicitly maintaining consistency between the model's predictions for old tasks and those made prior to learning a new task. 
For example, Learning-Without-Forgetting (LwF) \cite{LwF} method uses knowledge distillation~\cite{knowledge_distilation} to ensure that the model's predictions for previously learned tasks remain consistent with those before learning the new task. 
Distillation loss penalizes deviations in outputs for old tasks while learning new ones.

\subsubsection{Gradient Projection}
\hfill\\
Gradient projection~\cite{qiao2024gradientprojectioncontinualparameterefficient} minimizes catastrophic forgetting by constraining the gradients during the optimization/training.
The idea is to ensure that the updates for new task do not interfere with the learned parameters for previous task(s).
This is achieved by projecting gradients for the new task to a space that is less likely to conflict with 
gradients critical for previous tasks, this relies on geometric principles to identify such subspaces.
Gradient Episodic Memory (GEM), a gradient projection method stores a small subset of previous task data to use them to constraint gradient updates. 
It has been developed to alleviates forgetting, while allowing beneficial transfer of knowledge to previous tasks~\cite{GEM}.
GEM has been applied in reinforcement learning for robotics to learn new tasks while retaining knowledge on previous tasks~\cite{GEM}.
An extension of GEM called Average Gradient Episodic Memory (A-GEM) simplifies the gradient projection step by ensuring that the averaged gradients of stored samples from previous tasks are aligned with the new task gradient.
A-GEM has been tested in healthcare for disease classification across datasets and autonomous driving tasks~\cite{A-GEM, A-GEM2}.
In a slightly different approach, Orthogonal Gradient Descent (OGD)~\cite{OGD} projects the gradients for new task orthogonal to the previous task(s) gradients. To remove the need of storing gradients in the memory in OGD, more advanced techniques \cite{saha2021gradient,lin2022trgp,lin2022beyond} have been developed by extracting and storing important bases of the input subspace.

\subsubsection{Ensemble Methods}
\hfill\\
Ensemble-based continual learning methods integrate multiple models to improve adaptability and performance in dynamic environments. These approaches leverage the diversity of individual learners to mitigate catastrophic forgetting while maintaining overall model accuracy~\cite{CL_survey_ensemble}. 
Advantages of ensemble methods include enhanced robustness, improved generalization across tasks, and the ability to incorporate new knowledge without extensive retraining~\cite{learning_without_forgetting}. 
However, there are some challenges associated such as increased computational overhead, model management complexity, and potential difficulty in ensuring cohesive decision-making among diverse models~\cite{Ensemble_KNN}
Ensemble-based continual learning methods are broadly classified into two categories namely online ensembles and block ensembles.

Online ensemble methods incrementally update models without revisiting past data, ensuring adaptability to non-stationary distributions. 
Each incoming task or data batch may correspond to a lightweight model that is dynamically integrated into an ensemble.
Techniques such as online boosting or bagging are common examples, where models are added or re-weighted based on their performance on new data~\cite{bagging_and_boosting}. 
Online ensembles are particularly effective in scenarios with frequent distribution shifts, but their scalability and memory constraints remain a challenge in long-term learning ~\cite{bagging_and_boosting_book}.

In contrast, block ensemble methods operate in discrete phases, where each block corresponds to a specific task or data batch. 
Instead of incrementally updating a model, separate models or components are trained for each block and then combined into an ensemble. 
Methods such as expert networks associate individual blocks with specific tasks and use gating mechanisms to select the most relevant expert for predictions~\cite{expert_networks}. 
Although block ensembles are robust against forgetting, their computational and memory requirements grow with the number of tasks, limiting scalability.

\subsubsection{Meta Learning}
\hfill\\
Meta learning, often referred to as ``learning to learn", focuses on adapting quickly to new tasks by leveraging prior knowledge gained from previous tasks~\cite{Meta-learning0}. 
This paradigm is particularly useful in scenarios where data for a specific task is scarce, allowing models to generalize better from limited examples.
Meta-learning typically involves learning from a distribution of tasks. 
The idea is to capture the underlying patterns that can be generalized across different tasks. 
This can be formalized as learning a meta-model that optimizes performance on a variety of tasks, enabling rapid adaptation to new tasks~\cite{Meta-learning1}.
One common strategy in meta learning is to employ model-based methods, where the architecture of the model itself can adapt based on the learned representations. 
For instance, MAML (Model-Agnostic Meta-Learning) is a popular algorithm that trains a model such that it can achieve good performance on new tasks after only a few gradient updates~\cite{Meta-learning1}.
Another approach involves optimizing the learning algorithms themselves. 
This can include tuning hyperparameters or even learning optimization algorithms that are more effective for specific classes of tasks~\cite{Meta-learning2}.
Metric-based approaches utilize learned distance metrics to facilitate quick adaptation. 
For example, in few-shot learning scenarios, models like Prototypical Networks learn to compute distances between data points in an embedding space, enabling effective classification of new classes based on a few examples~\cite{Meta-learning3}.

Meta-learning has shown promise in various applications, including few-shot learning, where models are required to learn from a limited number of labeled examples~\cite{Meta-learning6, meta-learning7}. 
This has shown to be particularly beneficial in domains like healthcare and robotics, where obtaining labeled data can be costly and time-consuming~\cite{Meta-learning4}. 
Additionally, meta-learning techniques have been applied to improve the efficiency of reinforcement learning agents, allowing them to adapt more rapidly to new environments~\cite{Meta-RL}.
This includes a recent study on adapting RL to changing conditions in CERN's AWAKE project with the help of meta-RL~\cite{Meta-RL-CERN-DESY, Hirlaender:607018}. 
The study demonstrate meta-RL's ability in managing partially observable Markov decision processes with evolving hidden parameters prevalent in particle accelerators.

Offline Meta Learning involves training models on a fixed set of tasks or datasets before deployment. The model learns to generalize across these tasks, and once trained, it can quickly adapt to new tasks during inference without further learning from new data. This approach emphasizes the pre-training phase, where the model accumulates knowledge from a broad range of experiences.
Offline meta-learning typically relies on a well-defined distribution of tasks, allowing the model to learn generalizable strategies~\cite{Meta-learning6}.

Online Meta Learning refers to methods where the model learns continuously from a stream of tasks or experiences. 
This approach allows the model to update its knowledge dynamically as new data becomes available, facilitating rapid adaptation to new tasks with minimal retraining. 
The core idea is to leverage immediate feedback from newly encountered tasks to improve performance on future tasks.
Models can adjust their parameters in real-time based on incoming data, making them suitable for environments where tasks are non-stationary~\cite{Meta-RL}.
Typically, online meta-learning is employed in scenarios where the model has to learn from a limited number of examples for each new task~\cite{finn2019online,sow2024algorithm}.

\subsubsection{Online Continual learning}
\hfill\\
Unlike offline continual learning where each task has an offline dataset and the model can be updated using the dataset for multiple times, in online continual learning \cite{aljundi2019gradient} data arrives in streams and can only be trained once. 
This setup is particularly advantageous in dynamic environments with pure online data and low latency requirements. 
It seeks to update the ML models quickly and continuously using the current small amount of data, without forgetting the knowledge gained previously. 
The strategies to address forgetting in online continual learning follow a similar line as in offline continual learning, which implies that many offline continual learning approaches can be adapted to the online scenario. 
In the meanwhile, the online scenarios put a special focus on the learning effectiveness of ML models with a few data points.


\section{Deployable ML for Accelerators and Supporting Infrastructure}
\label{ch:deployableML}
\begin{figure}
    \centering
    \includegraphics[width=0.85\linewidth]{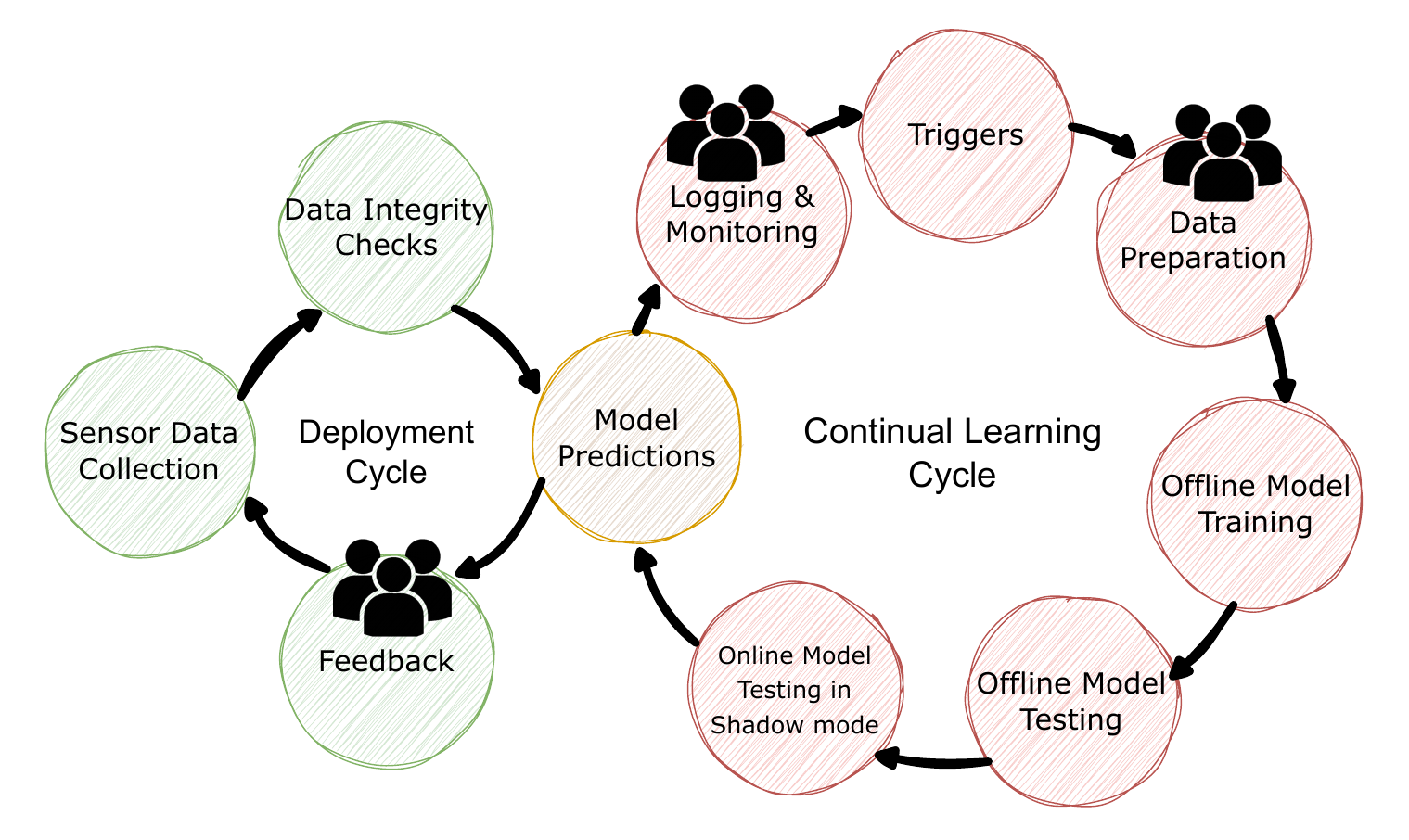}
    \caption{Continual learning pipeline with potential human in the loop in parallel to a typical deployment pipeline}
    \label{fig:CL_cycle}
\end{figure}
Traditional deployment of ML typically involves a static model that is trained on a fixed dataset before being deployed in a production environment. 
These deployment strategies typically perform periodic retraining of the ML models on new data that can lead to catastrophic forgetting.
Furthermore, the response to changes in data distribution can be slow, potentially degrading model performance until the next scheduled retraining cycle.
In contrast, continual learning facilitates a dynamic learning process where models continuously adapt to new data and operational conditions. 
This method allows for incremental updates without retraining from scratch, employing strategies such as fine-tuning and memory consolidation to retain prior knowledge.
Continuous monitoring of model performance enables rapid adaptation to shifts in data distribution, enhancing the resilience and relevance of ML applications. 
The continual learning framework thus offers a more robust and scalable solution for environments characterized by high variability and large data volumes, making it particularly suited for applications within particle accelerators.

\subsection{Continual Learning workflow}
Traditional deployment setups typically include a four step loop between data collection, integrity checks, model inference and decision making.
Adapting ML models via continual updates requires a parallel pipeline as shown in Figure~\ref{fig:CL_cycle}.
It includes data logging and continuous monitoring to trigger model adaptation when model degradation is detected, fine-tuning of the model, testing the new instance of the model in both offline and online (in shadow) mode before potentially replacing previous model.

Particle accelerators requires a significant amount of data storage, due to the amount of information that is accumulated over several years, across multiple runs. 
As such, many accelerator facilities only store a small amount of relevant data for future analysis, or save data at a lower rate. 
For facilities that do save comprehensive accelerator operation data at regular intervals, sorting through the data can become difficult. 
With recent surge in application of ML solutions in this space, many researchers have developed custom solutions to store desired data for training and evaluation of models.
For continual learning and model adaptation, depending on the selected approach, data logging becomes a necessity. 
However, it can be limited to certain time-frame and data samples can be filtered to avoid storing uninteresting samples. 
For example, in case of anomaly prediction models where normal samples are very similar in distribution, ML model adaptation may only need a small subset of normal data samples.

While the model is providing predictions, the monitoring module continuously observes all the input samples and model predictions to detect any drift in the data and model degradation. 
The selection of drift detection methods depends on underlying application. 
For example, for cases where labels become available immediately after data collection, predictive performance based drift detection can be very effective~\cite{error-based-drift-detection, average-accuracy-drift-detection, average-accuracy-drift-detection2, Concept_drift}.
However, in cases where decision making process influence the availability of labels, it becomes challenging. 
For example, in anomaly prediction applications in particle accelerators where beam is aborted when the model predicts an upcoming anomaly, it is challenging to know if the alarm was false because the beam has just been aborted.
In such cases, some proxy metrics are required to monitor predictive performance.
Window-based correlation methods that do not consider model predictions but monitor input data distributions to detect drifts can also be used.
When a drift is detected, model adaptation needs to be triggered. 
This step simply invokes a parallel process to start a training cycle. 
This training cycle may train a new model or fine-tune the model using existing weights and biases.

When model adaptation is triggered, it launches a parallel process responsible for data preparation, model re-training and testing.
The data preparation is entirely dependent on the model adaptation approach. For example, memory-based methods requires all the data including previous data distributions which would require storage of previously collected data. 
This can be improved up-to some extend by using a clever sampling technique to avoid storing all the data.
On the other side, regularization based adaptive methods does not require data from previous data distributions. 
The training dataset can be divided into three sets, namely training set, validation set, and test set. 
The test set in this case can be used to evaluate the ML model offline once trained/fine-tuned.
It is important to note that in many applications where we desire the model to retain previous knowledge, the model should be tested on data sets from previous distributions to ensure it maintain predictive performance.

If a new instance of the model provides better performance than previous version on offline datasets (including both previous distributions and new distribution), it should be tested online in shadow mode to make sure it provides better predictions on incoming data before replacing the old model.

\subsection{Supporting Infrastructure}

The successful implementation of continual learning in particle accelerators requires a comprehensive infrastructure that seamlessly integrates hardware, software, and organizational components. 
The infrastructure can be divided into two parts, namely hardware and software. 

Hardware requirements can be further modularized into two sets, one set for running online inference and feedback similar to traditional deployment cycle and second set of hardware to host continual learning processes.
The hardware for executing rapid inferences in real-time applications is dependent on the computational and latency needs for the application. 
If very low latency is needed, the hardware for inference may need to be deployed on the ``edge,'' i.e. close to the accelerator devices themselves. This could include ASIC/FPGA if latency requirements are extremely low (e.g. less than ms). 
For applications with more lenient timing restrictions but only modest compute requirements, GPU/CPU systems directly on the accelerator controls network may suffice. 
In the context of continual learning, to ensure that data can be effectively utilized for future model updates, a dedicated storage server is typically needed for logging and archiving operational data. 
Existing control systems may have a standard archiver, which may save data locally or on a separate network. 
Where the data is stored and how much historical and immediate data is used in the continual learning application will have an impact on where the actual continual learning process occurs (e.g. locally or on a separate network).

For applications with lightweight ML model and modest computational requirements, a local CPU/GPU machine can be hosted within accelerator network to facilitate continual learning cycle including monitoring, model re-training, and testing.
In contrast, for applications requiring more compute or access to larger data sets, interfacing with HPC systems such as local compute clusters may be needed. 
This does come with potential latency increases and considerations for how job scheduling is handled particularly in the context of large clusters with shared users. 
For use in online and continual learning applications, some HPC systems may have a reservation system or prioritized queue. 
Tools like Kubernetes~\cite{kubernetesGettingStarted} can help support long-running jobs on clusters, for applications like streaming data processing and continual learning. 
Both quasi-real-time inference (e.g. on the order of seconds) and interaction with the control system, as well as continual learning loops can be conducted on such HPC systems, provided a sufficient network bandwidth and speed is available.
For example, SLAC has connected its control system to its local HPC facility, the SLAC Shared Science Data Facility (S3DF), where compute intensive particle-in-cell simulations of the accelerator can be run from machine readings in quasi-real-time, suitable for use in a control room (e.g. minutes). 
Using the same hardware and software infrastructure, ML models are able to be deployed and put into a continual learning loop. 
Similarly, SLAC uses the National Energy Research Scientific Computing Center (NERSC) SuperFacility API \cite{NERSC_Superfacility_API} to enable photon science users to quickly analyze data and display results in the control room; this is now being expanded to include online retraining of ML models used in analysis. 
SLAC and LBNL are working on connecting online modeling / digital twin infrastructure to NERSC through the SuperFacility API. 
The US Department of Energy is planning to further enable these types of connections through its Integrated Research Infrastructure.

To support continual learning, software infrastructure is also needed. The sophistication of this may vary depending on whether edge compute, compute on the controls network, or separate HPC clusters are used. 
Tools such as MLFlow~\cite{mlflow} help with tracking and updating of ML models, and come with an associated model database. 
Tools such as Prefect~\cite{prefectPythonicModern} and AirFlow~\cite{airflow} can aid workflow orchestration, for example to keep track of and spawn processing associated with the ML monitoring and updating workflow automatically. 
Finally, tools like Kubernetes can help with orchestration of containerized applications, which is especially useful on shared HPC clusters. 
Some HPC facilities, such as NERSC, have custom APIs including for quasi-real time interaction.

Network requirements are critical for continual learning infrastructure. 
High-speed communication links are essential for facilitating data transfer between hardware components, as well as for the timely acquisition of sensor data. 
A low-latency network is vital for minimizing delays in data processing and model inference, ensuring that real-time applications can operate smoothly.
This requirement maybe trivial to deal with for continual learning on edge hardware or CPU/GPU machine inside accelerator network and close to the sensors. 
However, it is more complicated for continual learning on HPC.



\subsection{Selection of Continual Learning Method(s)}
ML model deployment use cases can broadly be classified into two categories, i.e., online deployment that provide ``real-time" decision support and offline as a surrogate.
For online deployed models, one of the main deciding factors for selection of continual learning method is latency. 
When drift causes the model to become obsolete, and the continual learning process is updating it to replace the old one, there is a period during which inferences can not be used for decision making. 
This is due to the fact that the predictions from the old model may no longer be reliable and should not be used for decision making.
This interruption can be referred to as ML downtime.
It is desired to minimize the ML downtime by enabling quick adaptation of the models in real-time use cases.
In contrast, for offline models, this requirement can be relaxed.
In this section, we revisit the applications of ML in particle accelerators as discussed in section~\ref{ch:background} and provide insights on which continual learning method(s) are most appropriate to handle the concept drifts.
This is mainly guided by the constraints on a given application and the properties of various continual learning methods as described in figure~\ref{fig:methods_table}.

\subsubsection{Anomaly Detection}
\hfill\\
Anomaly detection constitutes a significant domain of application within particle accelerators. 
This includes detection of gradually failing equipment as well as rapid deflection in beams.
ML models developed for anomaly detection or prediction are run online in real-time.   
Such models that provide timely predictions are essential for real-time decision making that are crucial for accelerator operation.
For example, outputs from an anomaly predictions model are used to abort the beam to avoid damage to the equipment by the deflected beam.
As such, rapid adaptation to distribution shift is critical to avoid ML downtime.

In anomaly prediction applications, as the accelerators are running continuously, there are multiple samples collected every second (depending on nature of machine and their frequency). 
For example, SNS at ORNL produces 60 samples of beam current time series waveform every seconds where each waveform contains about 100K measurements~\cite{Rajput_2024} that is equivalent to 6M points in one second. 
These data samples can be continuously collected and stored but require a large storage and significant computing power to process.
Machines that meet these computational demands are typically not integrated within the accelerator network.
As such, coupled with low latency requirements, model adaptation approaches that require data from previous tasks such as memory-based methods are less suitable in such cases.
It is also desirable to adapt the models with limited number of samples for new task, especially for supervised learning ML methods that require labels. 
If a continual learning method requires a large number of samples for new task, it artificially increases the ML downtime until a sufficient number of samples has been collected and the model can reach adequate performance.
This is because predictions from the old model may no longer be useful for decision making.
Another constraint that is critical when choosing a continual learning method is whether clear task boundaries are defined. 
As discussed above, particle accelerator data can have two types of distribution shifts, one caused by known changes and others due to unknown/non-measured factors. 
Clear task boundaries can be defined in cases when distribution shift is due to known changes in accelerator settings, however it is challenging in the later types of drifts.
As such, methods that may require clear task boundaries may also be less suitable in this use case.

This leaves us with three options, regularization-based methods, meta learning, and incremental learning. 
Each of these three methods suffers with some level of forgetting on previous tasks.
With unknown drifts in the data, it is generally acceptable to tolerate some level of forgetting on the previous data distributions. 
However, instead of using one of these methods in it's pure form, it is advisable to fuse them for this task. 
For example, meta-learning can be used to group the tasks based on known drifts. 
To handle unknown drifts within these groups, incremental regularization methods can be applied. 
This new hybrid approach can provide several benefits, the grouping leveraging meta learning will divide many beam settings into several groups based on their similarity to avoid small number of samples in base-learners. 
Single base-learners per group can be quickly adapted with regularization based methods in conjunction with incremental learning to reduce ML downtime.

\subsubsection{Optimization and Control}
\hfill\\
Optimization tasks in particle accelerators range from optimization of the individual sub-systems to optimization of the global settings for various equipment to achieve a desired stable beam delivery.
In addition, particle accelerators can greatly benefit from autonomous control methods. In fact, many facilities are aiming towards self-driving particle accelerators.
For autonomous ML operational cycle, continual learning is of great importance to maintain ML performance with changing conditions.

As discussed in section~\ref{ch:background}, many ML models have been used for optimization and control including BO, RL, and MPC.
Some of these approaches, such as BO, learn a new model from scratch at each run, whereas others, such as RL, and MPC may use a persistant ML model for the control policy and/or the value function. 
All the three algorithms, BO, RL, and MPC can use a learned system model to improve performance, and this model needs to be adapted over time as the machine changes.
In addition, RL, and MPC can also benefit from continual adaptation of the policy (controller) model with changing conditions.

When particle accelerators are restarted for a new run, ML models may need to be adapted from previous runs due to drifts in the data or setting up a new set of conditions (e.g. new beam energy).
In high power machines the models are run offline, or during a low power operation before the full power operation of the accelerator.
For systems that operate only in a few target beam configurations, some level of forgetting on previous data distribution is generally acceptable while adapting the models.
However, for machines that are put into many different operating conditions, it is desirable to retain some learning from previous operating conditions while also adapting to new ones.
In addition, for online optimization, to minimize ML downtime, it is desirable to chose continual learning method that has low latency and does not require large amount of data samples for the new task.
Moreover, task boundaries can be defined for distribution shifts due to known changes but not for the drifts due to unknown factors.

With these constraints, memory-based methods are suitable for optimization tasks with a clever sampling techniques such as priority replay~\cite{PER} or Hindsight experience replay~\cite{HER} to achieve relatively low latency.
It is important to note that traditional circular buffer used in RL will not be sufficient as the data samples from all the previous distributions should be retained in the buffer to limit forgetting. 
This is further complicated by the slow and gradual drifts from unknown sources where distribution boundaries can not be defined. 
As such, custom solutions are needed to effectively maintain samples to cover the entire area of the input space.
In addition, gradient projection can be a good choice as it can allow quick adaptation when small dataset is available for new task.

For online control tasks, meta-learning can be used to continuously adapt the agent/controller with drifting data.
Meta-RL~\cite{nagabandi2019learning} has already been explored in other fields, though for offline multi-task learning, it can be adapted for continual learning with minimal modifications.
In addition, for methods such as DRL, and DMPC that involve deep learning models, regularization based methods can be combined with online continual learning methods for continuous adaptation of the models as this is naturally fits into the RL/MPC framework.

\subsubsection{Surrogate Models and Data Driven Digital Twins}
\hfill\\
Surrogate models are typically trained offline on real data. 
These models can be used for several tasks including building of environments for offline training of optimization and control algorithms such as RL or MPC.
These models are often trained on historical data, however, new data distributions may need to be introduced from outside of the training distribution.
For example, a surrogate model trained on historical data may need to be updated periodically with new measured data to keep up with real drifts in the physical machine.
It is more important in data driven digital twin models that need to keep up with the physical system and continuously adapt to drifting system behavior.
Such model adaptations also occur offline and are more flexible than online model adaptation.

In these cases, no forgetting on previous tasks is desired to be able to reproduce previous results up-to a certain time horizon in the past and to be able to learn new distributions caused by time-varying drifts.
Dataset from previous data distributions are typically stored and are available.
As such, to maximize performance and limit forgetting, a memory-based approach should be preferred. 
However, if the data volume grows and cause significant challenges in using all the data for fine-tuning, a priority based replay can be used ~\cite{PER} to both limit the amount of data stored and make re-training faster.
In addition, as more and more drifted data distributions are added, the model may get saturated.
To address this, architecture based continual learning methods (such as progressive neural networks) can be combined with replay to add more capacity to the model.

In addition, these models can greatly benefit from the estimation of OOD uncertainty to inform the user when test data samples arrive from outside of the training distribution. The uncertainty estimation can also be used as a drift detection mechanism. 

\subsubsection{Virtual Diagnostics}
\hfill\\
Virtual diagnostic models are often run in real-time in sync with the accelerator operation to provide indirect measurements relevant to the beam.
If the predictions from virtual diagnostics model are used for real-time decision making, low latency is typically required for the model adaptation to reduce the ML downtime.
Whereas, if the virtual diagnostics model is used for offline analysis of the acquired data, some latency in model adaptation is usually acceptable.

Within a shorter temporal scope, where data drift due to non-measured factors remains minimal, it is feasible to define distribution boundaries. 
This is especially critical for machines that transition between various settings in a shorter time-span.
In such setups, retaining information on prior data distributions within a certain temporal horizon is advantageous, as the machine frequently reverts to previous settings. 
Conversely, over longer durations where data drifts due to unmeasured factors are evident, establishing distribution boundaries becomes challenging. 
In addition, some level of forgetting is acceptable on data distributions that are relatively older as these may have been compromised by unknown drifts.

Similar to real-time anomaly prediction, in this use case, hybrid continual learning approaches are most suitable.
A time-based priority data buffer can be maintained per beam setting instance to store the data within a fixed time horizon. 
This buffer can be utilized in memory-based continual learning methods to enable faster model adaptation with minimal forgetting on distributions that are recent in time.
The priority can be assigned to samples in the buffer based on given figure-of-merit and existing model's predictive error. 
In this approach, there is a trade-off between number of beam settings that can be included (and the size of buffers) and adaptation speed.
As the number of beam settings may grow, the model adaptation may slow down. 
In such cases, Ensemble-based methods can be combined to incrementally learn a model per beam setting that is fine-tuned from a base model that is aware of all the beam settings.
This approach is advantageous in learning new unseen beam settings faster with minimal number of samples as the model training leverages base model instead of training from scratch.
Another option is to use architecture based methods in place of ensemble based method to allow better scalability in setups with limited computing and memory resources.
In addition, models and data distribution that are relatively older and are known to have unknown drifts can be discarded as they may no longer be valid even if the accelerator transition to those exact settings.

\section{Conclusion and Outlook}
\label{ch:conclusion}
Applications of Machine Learning (ML) models in particle accelerators struggles significantly with data distribution drifts. 
As such, there is a huge gap between the number of ML solutions being developed and number of these solutions being used long-term in operation.
This domain can significantly benefit from continual learning methods.

In this paper, we have provided a comprehensive and insightful guide on development of adaptive ML models tailored for various applications within particle accelerators. 
We delved into the existing landscape, offering a brief overview of the existing ML applications. 
Moreover, we explored the challenges associated with deployment of these models in operational environment, specifically emphasizing the relevance to continual learning. 
Subsequently, we describe a detailed analysis of existing continual learning methodologies, along with their limitations and constraints pertaining to applications in fast-paced accelerator space.
We have provided a mapping between various application use-cases in accelerators and continual learning methods for the seamless automation of model adaptation in response to the time-varying drifts in the measurements whilst introducing novel hybrid approaches that hold the promise of optimal suitability and groundbreaking potential.
In addition, we have provided a guidance on the robust infrastructure requirements based on our experience from SLAC national laboratory and SNS accelerator.
With this paper, we have attempted to open up a new direction to manage data drifts and inspire more research efforts towards deployable continual learning for particle accelerators.

\section{Acknowledgement}
\label{sec:ack}

This work was partially supported by the DOE Office of Science, United States under Grant No. DE-SC0009915 (Office of Basic Energy Sciences, Scientific User Facilities program).
This manuscript has been authored by Jefferson Science Associates (JSA) operating the Thomas Jefferson National Accelerator Facility for the U.S. Department of Energy under Contract No. DE-AC05-06OR23177. 
Oak Ridge National Laboratory is operated by UT-Battelle, LLC, under contract DE-AC05-00OR22725. 
The contributions from Auralee Edelen are supported by SLAC National Accelerator Laboratory, under contract DE-AC02-76SF00515 for the US Department of Energy.
The US government retains, and the publisher, by accepting the article for publication, acknowledges that the US government retains a nonexclusive, paid-up, irrevocable, worldwide license to publish or reproduce the published form of this manuscript, or allow others to do so, for US government purposes. DOE will provide public access to these results of federally sponsored research in accordance with the DOE Public Access Plan (http://energy.gov/downloads/doe-public-access-plan)

\section*{references}
\bibliographystyle{iopart-num}
\bibliography{main}

\end{document}